\documentclass[letterpaper, 10 pt, conference]{ieeeconf}  
\IEEEoverridecommandlockouts        
\usepackage{cite}
\usepackage{amsmath,amssymb,amsfonts}
\usepackage{algorithmic}
\usepackage{graphicx}
\usepackage{textcomp}
\usepackage{xcolor}
\usepackage{multirow}
\usepackage[font=small]{caption}
\usepackage{subfigure}
\usepackage{subfig}

\title{\LARGE \bf
Intention Enhanced Diffusion Model for Multimodal Pedestrian Trajectory Prediction
}

\author{Yu Liu, Zhijie Liu, Xiao Ren, You-Fu Li, and He Kong
\thanks{This paper is to be presented at the 28th IEEE International Conference on Intelligent Transportation Systems (ITSC), 2025.}
\thanks{Yu Liu, Zhijie Liu, Xiao Ren, and He Kong are the Guangdong Provincial Key Laboratory of Fully Actuated System Control Theory and Technology, the Southern University of Science and Technology, Shenzhen 518055, China. Yu Liu is also with the Department of Mechanical Engineering, City University of Hong Kong, Hong Kong SAR, China. Emails: yuliu254-c@my.cityu.edu.hk; 12332642@mail.sustech.edu.cn; 12431359@mail.sustech.edu.cn; kongh@sustech.edu.cn. You-Fu Li is with the Department of Mechanical Engineering, City University of Hong Kong, Hong Kong SAR, China. Email: meyfli@cityu.edu.hk.
    }
}

\begin{document}
\maketitle
\thispagestyle{empty}
\pagestyle{empty}

\begin{abstract}

Predicting pedestrian motion trajectories is critical for path planning and motion control of autonomous vehicles. However, accurately forecasting crowd trajectories remains a challenging task due to the inherently multimodal and uncertain nature of human motion. Recent diffusion-based models have shown promising results in capturing the stochasticity of pedestrian behavior for trajectory prediction. However, few diffusion-based approaches explicitly incorporate the underlying motion intentions of pedestrians, which can limit the interpretability and precision of prediction models. In this work, we propose a diffusion-based multimodal trajectory prediction model that incorporates pedestrians’ motion intentions into the prediction framework. The motion intentions are decomposed into lateral and longitudinal components, and a pedestrian intention recognition module is introduced to enable the model to effectively capture these intentions. Furthermore, we adopt an efficient guidance mechanism that facilitates the generation of interpretable trajectories. The proposed framework is evaluated on two widely used human trajectory prediction benchmarks, ETH and UCY, on which it is compared against state-of-the-art methods. The experimental results demonstrate that our method achieves competitive performance.

\end{abstract}

\section{Introduction}
Pedestrian motion prediction is a critical capability for autonomous vehicles \cite{ref1}. Given the observed trajectories of pedestrians, accurate forecasting of their future paths is essential for ensuring safe and efficient operation. A key challenge in trajectory prediction lies in the inherently stochastic and non-deterministic nature of human motion, which is influenced by many factors such as social interactions and the surrounding environment. Within this context, multimodal trajectory prediction methods seek to generate interpretable probability distributions over plausible future paths, allowing the model to capture inherent uncertainty. 

Recent generative approaches have shown the ability to produce a diverse set of plausible path predictions. Among these, diffusion models \cite{ref38, ref39} have emerged as a prominent tool for pedestrian trajectory forecasting. However, a limitation of existing diffusion-based approaches is that they do not incorporate prior knowledge, such as pedestrian motion intention, which reduces explainability and undermines the prediction performance.

Motivated by the above observation, we propose an Intention-enhanced Diffusion model (IntDiff) for multi-modal pedestrian trajectory prediction in this paper. Our approach incorporates pedestrians’ motion intentions as a prior in the diffusion model, thereby rendering the prediction pipeline both more interpretable and more efficient. The main contributions of this work are as follows: 

(1) We propose a novel diffusion-based framework, IntDiff, for multimodal pedestrian trajectory prediction, which explicitly incorporates pedestrian motion intentions into the generative process, thereby enhancing the interpretability of the learned representations.
 
(2) A motion intention recognition module is introduced to estimate pedestrian intention along lateral and longitudinal axes, and is integrated with a classifier-free guidance strategy to condition the diffusion model, effectively steering the generation process toward intention-aware predictions.

(3) Comprehensive experiments are conducted on benchmark datasets, demonstrating that the proposed approach can achieve competitive performance compared to state-of-the-art methods in pedestrian trajectory forecasting.

\section{Related Work}
\subsection{Pedestrian Trajectory Prediction}

Early works on trajectory prediction proposed models based on physical motion constraints \cite{ref7}. However, these approaches often struggle to capture the complexity and variability of human motion in dynamic environments, prompting a shift toward data-driven learning-based methods. Recent deep learning approaches have framed trajectory forecasting as a sequential estimation task. For example, Long Short-Term Memory (LSTM) and Recurrent Neural Networks (RNN) have been employed to capture temporal dependencies in pedestrian motion \cite{ref8, ref9}. Given the critical role of social interactions in influencing pedestrians’ potential future trajectories, several approaches have adopted graph-based structures to explicitly model interactions between pedestrians \cite{ref11, ref12}. Additionally, environmental features such as visual maps and LiDAR point clouds are considered and integrated into the prediction pipeline \cite{ref18}.

Due to the inherent indeterminacy and multimodal nature of pedestrian motion, various methods have been developed to capture these characteristics in trajectory prediction. Generative Adversarial Networks (GANs) \cite{ref20, ref22} and Conditional Variational Autoencoders (CVAEs) \cite{ref24, ref25} introduce latent variables sampled from probability distributions to model the stochastic behavior of predicted trajectories. However, existing methods face inherent limitations, such as the generation of unrealistic trajectories, highlighting the need for more robust models.

\subsection{Multimodal Trajectory Prediction }
To enable interpretable trajectory predictions, some approaches incorporate anchor points as prior information to guide the generation of multimodal trajectories. For example, TNT \cite{ref27} estimate agents’ endpoints as prior conditions for predicting multimodal trajectories and ADAPT \cite{ref28} jointly predicts the trajectories of all agents in the scene with dynamic weight learning. Alternatively, some strategies \cite{ref29, ref30} select candidate prototype trajectories as anchor sets, where each prototype defines a motion pattern that waypoints should follow, subject to necessary refinement. \cite{ref46} proposes an LSTM-based model for interaction-aware, multi-modal motion prediction driven by maneuver recognition. GIMTP \cite{ref47} introduces a driving intention-specific feature fusion mechanism, enabling the adaptive integration of historical and future embeddings to improve intention recognition and trajectory prediction. Close in spirit to these models, we integrate pedestrian motion intentions in this work to better capture pedestrian motion uncertainty.   

\subsection{Denoising Diffusion Model}
Diffusion models \cite{ref38, ref39, ref45}, as a class of generative models, employ a parameterized Markov chain to model the gradual transition from a noise distribution to the target data distribution. These models have emerged as a prominent research focus, demonstrating their success across various domains, including image synthesis \cite{ref31}, video generation \cite{ref32}, and natural language processing (NLP) \cite{ref33}. In the context of motion prediction, MID \cite{ref34} approaches pedestrian trajectory prediction as a reverse process of motion indeterminacy diffusion. TRACE \cite{ref35} introduces a guided diffusion model that enables users to constrain trajectories by specifying target waypoints, speed, and social groups, while also incorporating the surrounding environmental context. SingularTrajectory \cite{ref36} introduces a diffusion-based universal trajectory prediction framework designed to reduce performance disparities across five distinct tasks by unifying diverse human dynamics representations. DICE \cite{ref37} presents a novel framework featuring an efficient sampling mechanism coupled with a scoring module to select the most plausible trajectories. Inspired by these developments, in this work, we explicitly categorize pedestrian motion intentions and incorporate them into the diffusion-based trajectory prediction framework.

\section{METHOD}
\subsection{Problem Definition}
The trajectory prediction task entails estimating pedestrians’ future positions within a scene based on their observed past movements. The model receives as input the 2D spatial coordinates of pedestrians over the observed time steps $t \in \{1,2,\dots,t_{\mathrm{obs}}\}$, denoted by $X = \{X^i_1, X^i_2, \dots, X^i_{t_{\mathrm{obs}}}\}$, where $X^i_t = (x^i_t, y^i_t)$ represents the position of the $i$th pedestrian at time $t$. Similarly, the ground truth trajectory over the future time period $t \in \{t_{\mathrm{obs}}+1, t_{\mathrm{obs}}+2, \dots, t_{\mathrm{pred}}\}$ is denoted as $Y = \{Y^i_{t_{\mathrm{obs}}+1}, Y^i_{t_{\mathrm{obs}}+2}, \dots, Y^i_{t_{\mathrm{pred}}}\}$, and $Y_t^i = (x_t^i, y_t^i)$ represents the ground truth position of the $i$th pedestrian at time $t$ in the future. The objective of this work is to predict the future trajectory $\hat{Y}^i$ of the $i$th pedestrian and to estimate its future positions $\hat{Y}^i_t = (\hat{x}^i_t, \hat{y}^i_t)$ at each time $t$.

\begin{figure*}
\centering
\includegraphics[scale= 0.55]{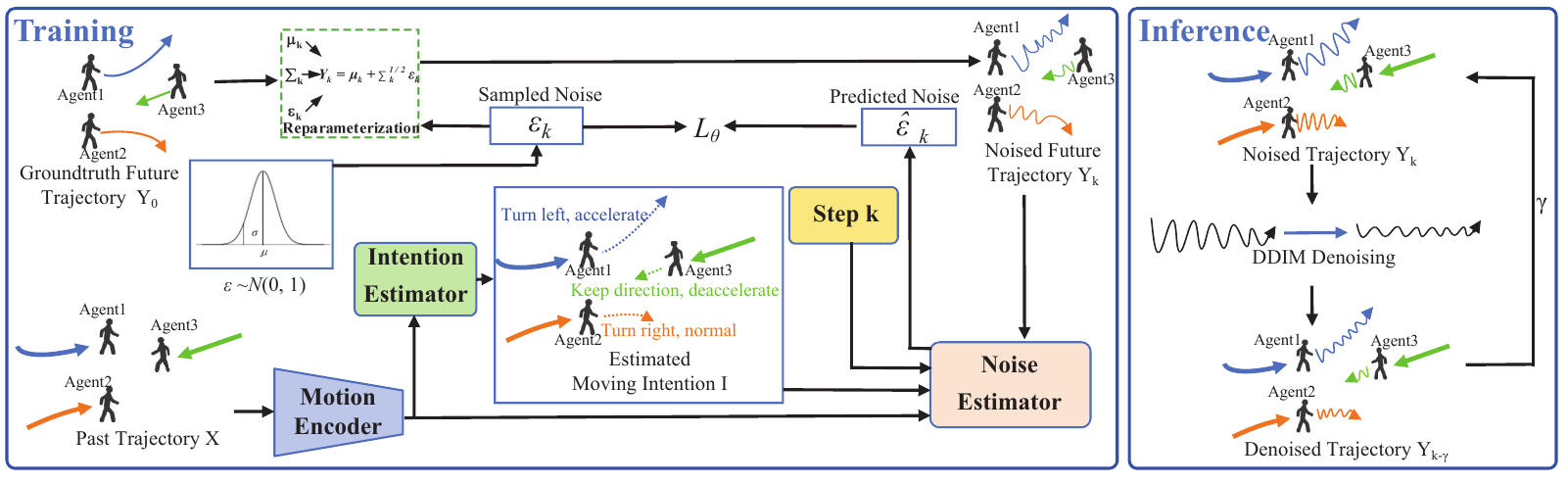}
\caption{The structure of the proposed IntDiff. During the training stage, pedestrian motion intentions are discriminated and integrated into the noise estimation process using the classifier-free guidance method. During inference, the DDIM sampling method is adopted to reduce computational cost.}
\label{figure structure2}
\vspace{-2em}
\end{figure*}

\subsection{Architecture Overview}
The overall architecture of IntDiff is illustrated in Figure~\ref{figure structure2}. It comprises three primary components: a motion encoder, which captures historical motion features of pedestrians; a motion intention estimator, based on a transformer architecture, which predicts future motion intentions; and a diffusion-based trajectory generator, which iteratively synthesizes future trajectories conditioned on both the encoded past motion and the estimated intentions. Detailed descriptions of each module are provided in the following subsections.

\subsection{Conditional Diffusion Model}
The diffusion model comprises two stages: a forward process and a reverse process. In the forward process, Gaussian noise is incrementally added to a sample drawn from the data distribution $Y_0$, which corresponds to the ground truth future trajectories. This process is repeated for $K$ steps following a predefined noise schedule, gradually transforming the sample into a standard Gaussian distribution. The forward process can be formally described as a Gaussian transition:
\begin{equation}
    \begin{array}{cc}
        q(Y_{1:K}{\mid}Y_{0} ) = \displaystyle \prod_{k=1}^{K} q(Y_{k}{\mid}Y_{k-1} ),
    \end{array}
\end{equation}
\begin{equation}
    q(Y_{k}{\mid}Y_{k-1} ) = \mathcal{N}(Y_{k}; \sqrt{1-\beta_{k}}Y_{k-1}, \beta_{k}I),
\end{equation}
where $\beta_k \in (0,1)$ denotes the rescaled variance schedule that controls the magnitude of noise added at each step. To reduce the computational cost during training, this process can be simplified using properties of Gaussian transitions:
\begin{equation}
    \begin{array}{cc}
        q(Y_{k}{\mid}Y_{0} ) = \mathcal{N}(Y_{k}; \sqrt{\bar{\alpha}_{k}}Y_{0}, (1-{\bar{\alpha}_{k}})I), 
    \end{array}
\end{equation}
\begin{equation}
    \begin{array}{cc}
        Y_{k} = \sqrt{\bar{\alpha}_{k}}Y_{0} + \sqrt{(1-{\bar{\alpha}_{k}})}\epsilon,\  \epsilon \sim \mathcal{N}(0, I),
    \end{array}
\end{equation}
where $\alpha_k = 1 - \beta_k$, $\bar{\alpha}_k = \prod_{i=1}^{k} \alpha_i$, and $\epsilon$ represents a noise vector sampled from a standard Gaussian distribution. As the diffusion step $k$ increases, the sample $Y_k$ gradually approaches the latent space represented by a standard normal distribution $\mathcal{N}(0, I)$.

For the reverse process of diffusion, a conditional parameterised Markov chain is initially designed to progressively denoise the input and generate future trajectories. However, this iterative scheme incurs substantial computational overhead and inference time. Moreover, the stochastic nature of the Markov chain may introduce deviations from the ground truth trajectory. To address these limitations, this work adopts the deterministic Denoising Diffusion Implicit Models (DDIM) sampling strategy \cite{ref41}, which eliminates stochastic terms during inference and allows for a reduced number of sampling steps. Under this approach, a sample $Y_{k-1}$ can be deterministically generated from $Y_k$ as follows:
\begin{equation}
\begin{aligned}
           Y_{k-1} = &\sqrt{\alpha_{k-1}} \left( \frac{Y_k - \sqrt{1 - \alpha_k} \, \epsilon_\theta(Y_k, X_k, I, k)}{\sqrt{\alpha_k}} \right) \\
           &  + \sqrt{1 - \alpha_{k-1}} \, \epsilon_\theta(Y_k,X_k, I, k),
\end{aligned}
\end{equation}
where $ \theta $ is the parameter of the diffusion neural network, $X_{k}$ is the corresponding observation sequence, and $I$ is the moving intention for feature. 

\subsection{Motion Encoder}
The motion encoder is designed to extract motion features from the observed trajectories of pedestrians while capturing their social interactions. These encoded features serve two purposes: they are input to the intention estimation module to predict future motion intentions, and they condition the diffusion model for trajectory generation. The encoding process can be formally expressed as:
\begin{equation}
F_{obs} =\mathcal{F}_\phi^{enc}(X) \in \mathbb{R}^{t \times d},
\end{equation}
where $X \in \mathbb{R}^{t \times d}$ denotes the observed historical trajectories of pedestrians, and $\mathcal{F}_\phi^{enc}$ represents the motion encoder parameterized by $\phi$. In this work, we adopt the encoder proposed in \cite{ref18}, which has demonstrated strong representation capabilities in capturing pedestrian motion patterns.

\subsection{Intention Estimation }
\subsubsection{Pedestrian Moving Intention }
The motion intention $I = \left\{I_{lo}, I_{la} \right\}$ of a pedestrian is defined from two perspectives: longitudinal $I_{lo}$ and lateral $I_{la}$. The lateral intention $I_{la} \in \left\{Lt, Rt, Kd \right\}$ represents turning left, turning right, or keeping direction along the vertical axis. Since explicit annotations of pedestrian intentions are not available in existing datasets, we infer them based on observed trajectories. Specifically, lateral intention is determined by the lateral velocity $v_{la}$, which is computed as the first-order derivative of the trajectory, and defined as follows:
\begin{equation}
I_{la} = 
\begin{cases}
\text{Lt} & \text{if } v_{la} > v_{lt} \\
\text{Rt} & \text{if } v_{la} < v_{rt} \\
\text{Kd} & \text{other},
\end{cases}
\end{equation}
where $v_{lt}$ and $v_{rt}$ represent the velocity thresholds for left turn and right turn, respectively.

For longitudinal intention, it is categorized into three patterns: accelerating, decelerating, and normal, $ I_{lo} \in \left\{Acc, Dec, Nor \right\} $, which describe motion patterns along the horizontal direction. Similarly, longitudinal intention is derived from pedestrian trajectories. Instead of relying on velocity, it is classified based on acceleration, the second-order derivative of position, and is defined as follows:
\begin{equation}
I_{lo} = 
\begin{cases}
\text{Acc} & \text{if } a_{lo} > a_{acc} \\
\text{Dec} & \text{if } a_{lo} < a_{dec} \\
\text{Nor} & \text{other},
\end{cases}
\end{equation}
where $ a_{acc} $ and $ a_{dec} $ are the acceleration thresholds for longitudinal acceleration and deceleration, respectively.

\subsubsection{Intention Prediction Module }
The intention prediction network infers pedestrians’ future motion intentions based on their historical motion features.
\begin{equation}
\hat{I} =\mathcal{F}_\psi^{int}(F_{obs}) \in \mathbb{R}^{t \times 2},
\end{equation}
where $\hat{I}$ denotes the estimated motion intention and $\mathcal{F}_\psi^{int}$ represents the intention prediction module parameterized by $\psi$. Here $\mathcal{F}_\psi^{int}$ is implemented using a transformer block.

\subsection{Intention Guidance}
One challenge in intention-guided generation is balancing the influence of conditioned motion intention. To address this, we employ classifier-free guidance \cite{ref42}, first computing an unconditional prediction and then adjusting it with the conditioned prediction. This methods leverages pedestrians’ motion intentions to guide the denoising process.
\begin{equation}
{\varepsilon}_\theta(Y_k, X_k, I, k) =  w \cdot \hat{\varepsilon}_\theta(Y_k,  X_k, I, k) +  (1 - w) \cdot \hat{\varepsilon}_\theta(Y_k, \varnothing, k ),
\end{equation}
where $\varnothing$ denotes the embedding corresponding to an empty condition, $w$ is a guidance scale parameter that regulates the influence of motion intention to avoid over-dependence, and $\hat{\varepsilon}_{\theta}$ represents a Transformer-based network employed to predict the noise.

\subsection{Training Optimization}

The training of the proposed method consists of two parts. The loss for the intention prediction module is formulated using the Mean Squared Error (MSE), which directly establishes a mapping between past motion features and future motion intentions, and is defined as follows:
\begin{equation}
L_{intent} = MSE(I, \hat{I}).
\end{equation}

For the trajectory prediction loss, we follow the Evidence Lower Bound (ELBO) maximization approach  \cite{ref39}:
\begin{equation}
\hspace{-0.3cm}
    \begin{aligned}
L_{\theta} &= \mathbb{E}_q  \Bigl[ \sum_{k=2}^{K} D_{KL} \left( q(Y_k \mid Y_{k-1}, Y_0) \,\|\, p_{\theta}(Y_{k-1} \mid Y_k, f) \right) \\
           &- \log p_{\theta}(Y_0 \mid Y_1, f) \Bigr] ,
    \end{aligned}
\end{equation}
where $f$ is the prior condition. The above cost function can be reformulated in the following form in an equivalent noise estimation problem:
\begin{equation}
\mathcal{L}_{\theta}(\theta) = \mathbb{E}_{\epsilon_{(0)}, Y_{(0)}, k} \left\| \epsilon_{(0)} - \hat{\epsilon}_{(\theta)}(Y_{(k)}, k, X_k, I) \right\|^2.
\end{equation}
Then, the loss is a weighted combination of all terms:
\begin{equation}
\mathcal{L} = \frac{1}{N} \left[\sum_{i=1}^{N} \alpha \mathcal{L}^{i}_{intent} + \beta\mathcal{L}^{i}_{\theta}\right],
\end{equation}
where $\alpha$ and $\beta$ are empirical weights used to balance the two loss components, and $N$ denotes the number of pedestrian trajectories in the training batch.

\section{Experiments and Results Analysis}

\subsection{Dataset}
In this study, the proposed model is trained and evaluated on two public pedestrian trajectory datasets: ETH \cite{ref43} and UCY \cite{ref44}. The ETH dataset comprises two scenes, ETH and HOTEL, while the UCY dataset includes three scenes: ZARA1, ZARA2, and UNIV. All scenes provide pedestrian positions in world coordinates, with results reported in meters at 2.5 Hz. We adopt the standard training and evaluation protocol established in prior works, where each sequence of 20 frames (8 seconds) is divided into two segments: the first 8 frames (3.2 seconds) are used as observed trajectory input, and the subsequent 12 frames (4.8 seconds) serve as the ground truth for future trajectory prediction.

\subsection{Evaluation Metrics}
To evaluate the performance of the proposed method, two standard metrics are employed: Average Displacement Error (ADE) and Final Displacement Error (FDE). ADE measures the average Euclidean distance between the predicted and ground truth trajectories over all prediction time steps, while FDE quantifies the Euclidean distance between the predicted final position and the corresponding ground truth.
\begin{equation}
    ADE=\frac{\sum_{n=0}^N \sum_{t=t_{obs}+1}^{t_{pred}}{\parallel{\hat{p}_t^n-p_t^n}} \parallel ^2}{N \times (t_{pred}-t_{obs})},
\end{equation}
\begin{equation}
    FDE=\frac{\sum_{n=0}^N \parallel {{\hat{p}_t^n-p_t^n}} \parallel ^2}{N}, \text{ } t=t_{pred},
\end{equation}
where $N$ is the total number of trajectories;  $t_{\mathrm{obs}}$ is the time frame of observed paths, while $t_{\mathrm{pred}}$ is the time frame of predicted trajectories; $\hat{p}_t^n$ and$p_t^n$ are the predicted trajectory point and ground truth trajectory point for the $n$-th trajectory at time frame $t$, respectively.

\subsection{Implement Details}
The network is implemented using the PyTorch framework. The noise estimation network in the diffusion model employs 4 transformer layers, each with a dimension of 512 and 8 attention heads. In the pedestrian intention estimator block, features are mapped to a dimension of 256, with 4 transformer layers and 4 attention heads. The velocity thresholds are empirically set to 0.2 and -0.2 for $v_{lt}$ and $v_{rt}$, respectively. $a_{acc}$ and $a_{dec}$ are set to 0.5  and -0.5, respectively. The guidance scale $w$ is set to 0.9. $\alpha$ is set to 1 and $\beta$ is set to 0.5. We use $K = 100$ diffusion steps in the forward process and adopt a stride of 20 with the DDIM sampling method in the inference process. The training batch size is set to 256, and the model is optimized using Adam with a learning rate of 0.001. All experiments are conducted on a system featuring Nvidia RTX 4090 GPUs and an Intel Xeon 4210R CPU.

\subsection{Quantitative Evaluation}
\begin{table*}[t]
\centering
\caption{ Quantitative ADE/FDE results on the ETH/UCY benchmark. \textbf{Bold} numbers indicate the best performance, and \underline{underlined} numbers represent the second-best results. * represents the results are referred to \cite{ref37}. }
\fontsize{9}{11}\selectfont
\label{result} 
\begin{tabular}{l||ccccc|c}
\hline
\textbf{Method} & ETH & Hotel & Univ & Zara1 & Zara2 & Avg \\
\hline \hline
SocialGAN \cite{ref22} & 0.81/1.52 & 0.72/1.61 & 0.60/1.26 & 0.34/0.69 & 0.42/0.84 & 0.58/1.18  \\ 
\hline
Social-STGCNN \cite{ref10} & 0.64/1.11 & 0.49/0.85 & 0.44/0.79 & 0.34/0.53 & 0.30/0.48 & 0.44/0.75 \\
\hline
Trajectron++* \cite{ref18} & 0.67/{1.18} & 0.18/0.28 & 0.30/0.54 & 0.25/0.41 & 0.18/0.32 & 0.32/0.55 \\
\hline
GroupNet \cite{ref40} & 0.46/0.73 & \textbf{0.15}/0.25 & 0.26/0.49 & 0.21/0.39 & {0.17}/0.33 & {0.25}/0.44 \\
\hline
MID* \cite{ref34} & 0.54/0.82 & 0.20/0.31 & 0.30/0.51 & {0.27}/0.46 & 0.20/0.37 & 0.30/0.51 \\
\hline
DICE \cite{ref37} & \textbf{0.24}/\textbf{0.34} & 0.18/\textbf{0.23} & 0.52/0.61 & 0.24/\textbf{0.37} & 0.20/\textbf{0.30} & 0.26/\textbf{0.35} \\
\hline
IntDiff (ours) & \underline{0.41}/\underline{0.71} & \textbf{0.15}/\underline{0.24} & \textbf{0.22}/\textbf{0.43} & \textbf{0.20}/\textbf{0.37} & \textbf{0.15}/\underline{0.31} & \textbf{0.23}/\underline{0.41} \\
\hline
\end{tabular}
\vspace{-1em}
\end{table*}

\begin{table}[h]
\centering
\caption{Ablation study on diffusion step ${K}$}
\setlength{\tabcolsep}{5pt}  
\fontsize{8}{10}\selectfont
\label{Ablation_diffstep}
\begin{tabular}{cccccc}
\hline
 & ${K}=20$ & ${K}=50$ & ${K}=100$ & ${K}=150$ & ${K}=200$\\
\hline
ETH     & 0.44/0.72 & \textbf{0.40}/\textbf{0.71} & 0.41/\textbf{0.71} & 0.41/0.74  & 0.45/0.81 \\
HOTEL   & 0.20/0.27 & 0.16/0.26 & \textbf{0.15}/\textbf{0.24} & \textbf{0.15}/0.25  & 0.21/0.30 \\
UNIV    & 0.26/0.48 & \textbf{0.21}/0.44 & 0.22/0.43 & 0.22/\textbf{0.42}  & 0.23/0.46 \\
ZARA1   & 0.27/0.39 & 0.22/0.40 & \textbf{0.20}/0.37 & 0.21/\textbf{0.36}  & 0.24/0.40 \\
ZARA2   & 0.17/0.35 & 0.18/\textbf{0.31} & \textbf{0.15}/\textbf{0.31} & \textbf{0.15}/0.32  & 0.19/0.33 \\
\hline
AVG     & 0.27/0.44 & 0.23/0.43 & \textbf{0.23}/ \textbf{0.41} & \textbf{0.23}/0.42  & 0.26/0.46\\
\hline
\end{tabular}
\vspace{-1em} 
\end{table}

\begin{table}
\centering
\renewcommand{\arraystretch}{0.9}
\setlength{\tabcolsep}{9pt}  
\caption{Ablation study on motion intention}
\fontsize{8}{10}\selectfont
\label{Ablation_intention}
\begin{tabular}{ccccc}
\hline
 & Base & w/o $I$ & w/o $I_{la}$ & w/o $I_{lo}$ \\
\hline
ETH     & \textbf{0.41}/0.71 & 0.45/0.75 & 0.42/0.74   & 0.43/\textbf{0.70}\\
HOTEL   & \textbf{0.15}/\textbf{0.24} & 0.21/0.30 & \textbf{0.15}/0.27   & 0.20/0.27\\
UNIV    & \textbf{0.22}/\textbf{0.43} & 0.30/0.50 & 0.25/0.52   & 0.26/0.46\\
ZARA1   & \textbf{0.20}/0.37 & 0.27/0.43 & 0.21/0.40   & 0.25/\textbf{0.36}\\
ZARA2   & \textbf{0.15}/0.31 & 0.20/\textbf{0.30} & 0.19/0.34   & \textbf{0.15}/0.34\\
\hline
AVG     & \textbf{0.23}/\textbf{0.41} & 0.29/0.46 & 0.24/0.46   & 0.26/0.43\\
\hline
\end{tabular}
\vspace{-1em}
\end{table}

\begin{figure*}[t]
  \centering
  \includegraphics[width=0.85\linewidth]{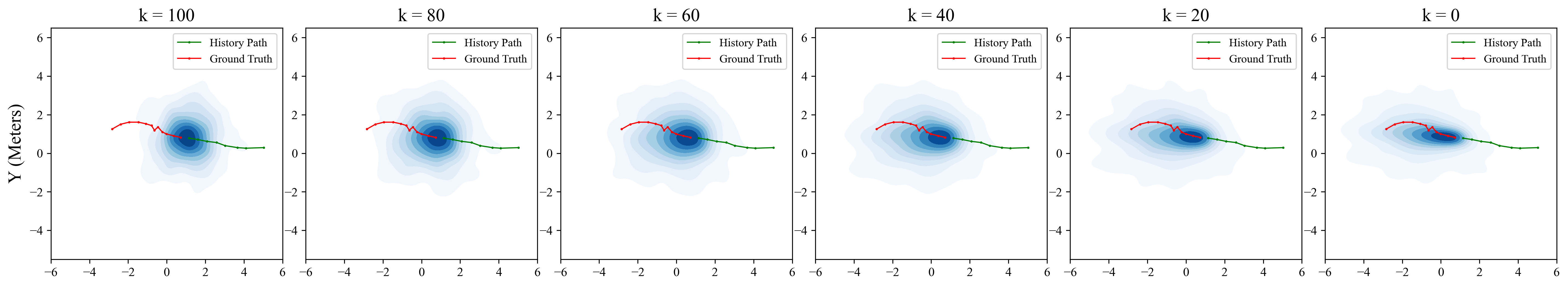}
  \vspace{-0.30em}
  \hspace{-0.5em}
  \includegraphics[width=0.85\linewidth]{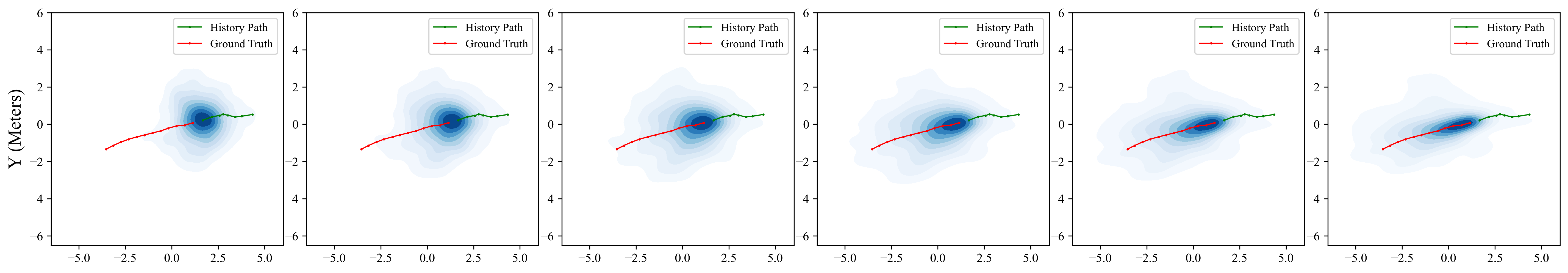}
  \vspace{-0.30em}
  \hspace{-0.1em}
  \includegraphics[width=0.85\linewidth]{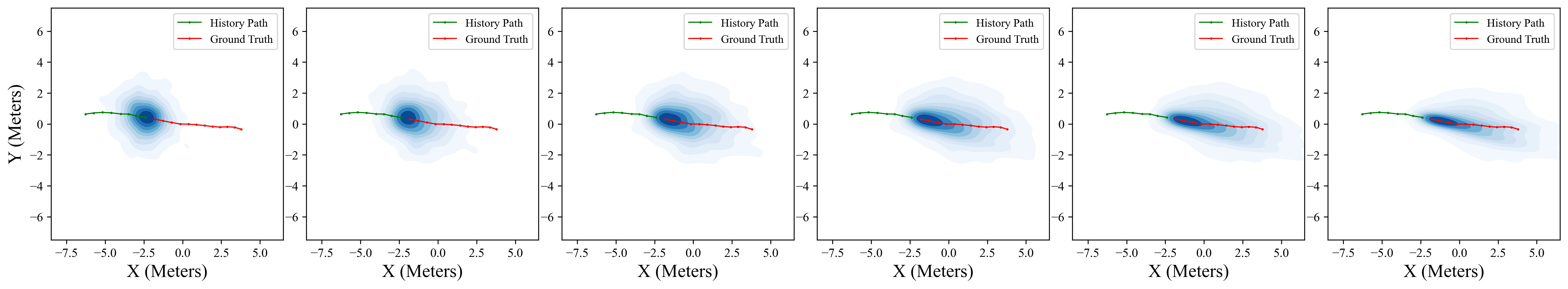}
  \vspace{-0.20em} 
  \caption{Visualization of the trajectory generation process across three scenes. For each example, generated samples are recorded and their densities are visualized. Observed trajectories are shown as green lines, while red lines indicate the ground truth. X and Y denote the predicted latitude and longitude positions, respectively.}
  \label{Visual1}
  \vspace{-0.5em}  
\end{figure*}

\begin{table}[h]
\centering
\caption{Ablation study on classify guidance scale ${w}$}
\setlength{\tabcolsep}{5pt}  
\fontsize{8}{10}\selectfont
\label{Ablation_omega}
\begin{tabular}{cccccc}
\hline
 & ${w}=0$ & ${w}=0.25$ & ${w}=0.75$ & ${w}=0.90$ & ${w}=1$\\
\hline
ETH     & \textbf{0.40}/0.72 & 0.43/0.73 & 0.45/0.72 & 0.41/\textbf{0.71}  & 0.43/\textbf{0.71} \\
HOTEL   & 0.17/0.27 & 0.18/0.28 & 0.16/0.26 & \textbf{0.15}/\textbf{0.24}  & \textbf{0.15}/0.25 \\
UNIV    & 0.25/0.47 & \textbf{0.21}/0.48 & 0.23/0.44 & 0.22/\textbf{0.43}  & 0.22/0.45 \\
ZARA1   & 0.21/0.41 & 0.26/\textbf{0.36} & 0.22/0.37 & {0.20}/0.37  & \textbf{0.19}/0.39 \\
ZARA2   & 0.20/\textbf{0.30} & 0.18/0.31 & \textbf{0.15}/\textbf{0.30} & \textbf{0.15}/0.31  & 0.17/0.33 \\
\hline
AVG     & 0.25/0.44 & 0.25/0.43 & 0.24/0.42 & \textbf{0.23}/\textbf{0.41} & \textbf{0.23}/0.43\\
\hline
\end{tabular}
\vspace{-2em}
\end{table}

The proposed approach is compared against a wide range of recent methods, including SocialGAN \cite{ref22}, Social-STGCNN \cite{ref10}, Trajectron++ \cite{ref18}, GroupNet \cite{ref40}, MID \cite{ref34} and DICE \cite{ref37}. The quantitative evaluation results on UCY/ETH for the proposed approach compared to existing methods are presented in Table \ref{result}. It can be observed that, despite variations across subsets, our method achieves competitive performance in most cases. Specifically, on the Univ scenario, the proposed method achieves the overall best performance by a significant margin, outperforming GroupNet by approximately 15\% and 12\% on ADE and FDE, respectively. An 11\% reduction in ADE is also observed on the Zara2 subset. On average across all subsets, our model achieves the best ADE, with an improvement of up to 8\% compared to GroupNet, while ranking second in terms of FDE. To further investigate the contribution of each component, ablation studies are conducted in the sequel.

\subsection{Visualization}
The reverse process of the diffusion model on a subset of the ETH dataset is illustrated in Figure \ref{Visual1}. In this experiment, the total number of diffusion steps is set to 100, with a sampling stride $\gamma$ of 20 during the DDIM sampling process, resulting in five sampling steps. For each frame, 200 samples are generated and recorded. As shown in the figure, where color intensity reflects the sample density, the initially diverse distribution of predicted points gradually converges toward the ground-truth trajectories as it follows the overall motion trend and eventually stabilizes.     

\subsection{Ablation Studies}
In the proposed method, an intention-enhanced diffusion model is utilised to generate future trajectories, where the number of steps controls denoising. To evaluate its impact, we vary the values of the diffusion step $K$ from 20 to 200, and the results are shown in Table \ref{Ablation_diffstep}. It can be observed that generally, when $K$ reaches 100 steps, both ADE and FDE reach their minimum. With smaller diffusion steps, the model may fail to capture the detailed transitions of noise between each step, while with larger diffusion steps, the subtle variations between steps become harder to discriminate due to the limited capacity of the neural network.

In this work, pedestrians' motion intentions $I$ are split into lateral $I_{la}$ and longitudinal $I_{lo}$ perspectives. To verify the impact of each component, the next ablation study investigates the effects of $I$, $I_{la}$, and $I_{lo}$ respectively. The results are summarized in Table \ref{Ablation_intention}. It can be observed that the loss of any intention vector leads to performance degradation, which highlights the significance of intention as the prior conditional distribution. Similarly, an ablation study on the impact of intentions guidance scale $w$ is conducted as illustrated in Table \ref{Ablation_omega}. When $w$ reaches 1, the prediction is dominated by the conditional model. On the contrary, prediction will be guided only by the unconditional model when $w$ is 0. We set the dynamic balance factor $w$ to 0.9, as it yields better overall performance across all datasets.   

\section{Conclusion}
We propose IntDiff, an intention-enhanced diffusion model for multi-modal pedestrian trajectory prediction, incorporating pedestrians' motion intention. The proposed approach consists of a motion intention recognition network for estimating pedestrians' future intentions as a prior distribution for the diffusion model. Trained with the classifier-free guidance method, our approach is capable of generating more interpretable trajectories. Comparative experimental results demonstrate the effectiveness and robustness of our model compared to existing methods. To better mitigate unstable performance across different scenes, environmental and social context will be incorporated into the prediction process as future work.

\section*{ACKNOWLEDGMENT}
This work was supported by the National Key R\&D Program of China under Grant No. 2024YFB4710902, the National Natural Science Foundation of China under Grant No. U24A20265, the Shenzhen Science and Technology Program under Grant No. KQTD20221101093557010, the Guangdong Science and Technology Program under Grant No. 2024B1212010002.
\bibliographystyle{IEEEtran}
\bibliography{refs}

\end{document}